\documentclass[11pt]{article}
\usepackage[preprint]{acl}
\usepackage{times}
\usepackage{latexsym}

\usepackage{amsmath,amsfonts,bm}

\def\eqref#1{equation~\ref{#1}}

\def\1{\bm{1}}

\DeclareMathAlphabet{\mathsfit}{\encodingdefault}{\sfdefault}{m}{sl}
\SetMathAlphabet{\mathsfit}{bold}{\encodingdefault}{\sfdefault}{bx}{n}

\usepackage{url}
\usepackage{graphicx}
\usepackage{float}
\usepackage{subcaption}
\usepackage{booktabs}
\usepackage{longtable}
\usepackage{amsmath}
\usepackage{amssymb}
\usepackage{xcolor}
\usepackage{microtype}

\title{Can LLM User-Proxies Stand In for Humans in Reliance Measurement? Replicating Rel-A.I. with Simulated Users}

\author{Shuai Wang \qquad Rui Hu \\
  Zhejiang University}

\begin{document}
\raggedbottom

\maketitle

\begin{abstract}
Can language-model user proxies reproduce human reliance on an AI answer across contexts? Published simulated-user studies do not establish validity for this partial-answer decision, and Rel-A.I.'s released human record omits trial-level choices. We audit its 90 printed questions and 15 domain-by-marker reliance rates using sampled decisions and counterbalanced A/B probabilities from open models through 72B parameters. A matched crossover varies the printed agent specialty within each question, separating question content, label averages, and their interaction. Mistral Small's strong-marker math-minus-law label marginal is $-0.4$ percentage points, yet matching the agent label to the question domain raises reliance by 13.3 points on average across the five domains. The best A/B proxy has 22.5-point absolute-rate error and 8.3-point domain-effect error against the human table. A two-sample independent-binomial human replication has 3.5-point expected domain-effect error under the stated working model. The interaction is model dependent, and inference layout changes identical-prompt probabilities. This single-study audit does not establish proxy validity for human effect sizes or general directional screening.

\end{abstract}

\section{Introduction}
\label{sec:intro}

\begin{figure}[H]
  \centering
  \includegraphics[width=\columnwidth]{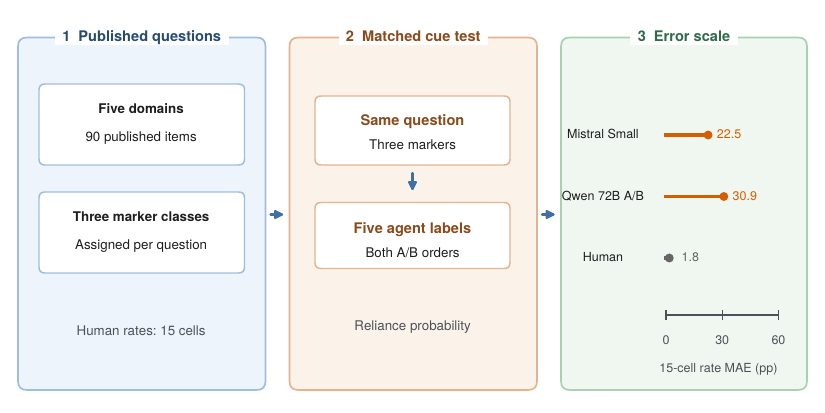}
  \caption{\textbf{The audit pairs a published human table with matched proxy questions.} The left panel shows the printed Rel-A.I. questions and epistemic-marker classes; item-to-marker assignments are reconstructed. The middle panel shows the matched question, marker, agent-label, and option-order crossover. Lollipops at right compare 15-cell rate errors for two A/B proxies with a plug-in independent human resample.}
  \label{fig:teaser}
\end{figure}

People often decide whether to use an AI answer before they can verify it. The wording of a partial answer and the domain of the question can change that decision. Rel-A.I. measures reliance by showing people the beginning of an agent's answer and asking whether they will use the completed answer or look it up themselves \citep{zhou2025rel}. Its domain study reports rates across five subjects and three classes of certainty language. Repeating the study for every interface or model revision would require new participants. A language-model proxy could be useful if it preserves the human changes across conditions.

Validation requires more than a plausible single choice. A screening instrument must preserve the ordering of conditions, while an effect-size substitute must also reproduce the distance between them and the baseline reliance level. Simulated users can misestimate human task success \citep{seshadri2026lost}, and survey simulations can distort between-group differences \citep{chen2026synthetic}. Neither finding establishes how a simulated respondent behaves in Rel-A.I.'s partial-answer task. Its printed questions and aggregate human table provide a direct behavioral reference, although the original marker assignments and participant-level decisions are unavailable. Those missing records limit uncertainty analysis and exact replay.

We ask whether a proxy preserves reliance levels and changes across contexts well enough to prioritize conditions for a human study. A constant decision can appear accurate on small domain effects, while a preference for mathematical subjects can match many correlated pairwise signs. The missing experimental distinction is whether a domain shift follows the agent's stated specialty or the questions assigned to that specialty. Figure~\ref{fig:teaser} shows our matched test: the same question appears with all five agent-domain labels, each marker class, and both option orders. We retain the original sampled replay and a separate counterbalanced probability protocol to place this test against the published human table.

Our evaluation compares every published domain-by-marker rate, all within-marker domain pairs, and strong-minus-weak marker effects. Absolute rates describe how often someone would accept an answer in a condition. Domain contrasts indicate which subject might merit further testing; marker contrasts measure sensitivity to expressed certainty. A model can rank conditions while placing nearly every decision at the floor or ceiling. We therefore compare effect errors with a zero-domain-effect rule and a same-size human resample under a stated working model. Human-table-fitted marker and STEM rules are descriptive checks on the score, not deployable competitors. The human reference lacks participant records, so the resampling reference cannot reproduce its within-person covariance.

The lowest-error A/B proxy, Mistral Small, has 8.3 percentage points of domain-effect error. A two-sample independent-binomial human replication has 3.5 points of expected error under the stated working model. In the matched crossover, Mistral Small's strong-marker mathematics-minus-law label marginal is near zero, but a label matching the question domain raises reliance by 13.3 points on average. Thus a marginal label comparison can hide a substantial question-by-label interaction. Few human domain pairs resolve under the discrete-binomial working approximation, and the resolved set changes with assumptions about within-person correlation. This table does not establish a general directional screening ability.

This paper contributes a reconstruction that keeps published and inferred materials separate, an audit of the complete released domain table, and a within-question label crossover across three proxies. The crossover identifies a question-dependent effect of the printed agent-domain label on proxy decisions while leaving the model's own question accuracy unmeasured. The aggregate human record cannot determine within-person covariance or a matched greeting effect. The resulting conclusion is limited to this task and these protocols: the tested proxies do not support replacing human reliance measurements.

\section{Related Work}
\label{sec:related}

\paragraph{Behavioral reliance.} Rel-A.I. records a decision to defer to an incomplete agent answer and varies greetings, prior confidence, and question domain \citep{zhou2025rel}. Earlier human experiments by its authors examined how certainty markers and model reluctance to express uncertainty affect reliance \citep{zhou2024unreliable}. The reported Rel-A.I. domain rates provide the external target here. Self-reported trust instruments, including the Trust-In-LLMs Index, measure a related construct \citep{deduro2025trust}. We evaluate the observed rely-or-look-up choice and make no claim about a proxy's internal trust.

\paragraph{Simulated respondents.} Demographically conditioned ``silicon samples'' can approximate response distributions for survey subgroups \citep{argyle2022one}, and interview-grounded language-model agents can reproduce some survey responses and experimental outcomes \citep{park2026agents}. Simulated users in agentic evaluations can misestimate human task success and vary across simulator models \citep{seshadri2026lost}. Interactive literature-search agents are likewise evaluated against a fixed LLM user simulator that sees the hidden gold papers, with leak checks and a small human study because simulated feedback cannot fully replace users \citep{li2026paperpilot}. A cross-domain survey audit also finds inflated differences between demographic segments \citep{chen2026synthetic}. These studies make the target estimand central to validation. We test a behavioral rely-or-look-up decision against published human condition effects, including absolute rates and response-interface sensitivity.

\paragraph{Response interfaces.} Multiple-choice selection can shift when answer positions or option identifiers change \citep{zheng2024selectors}. Our A/B protocol balances option order and measures the residual difference between orders. The resulting rates remain conditional on the two labels, which is why we compare them with the sampled-response protocol as a separate measurement rather than attributing a difference to model psychology.

\section{Method}
\label{sec:method}

\subsection{Reference and estimands}
\label{sec:prelim}

Rel-A.I. asks a person whether to use an agent's hidden completed answer or seek the answer independently \citep{zhou2025rel}. Its domain study reports a reliance rate for each combination of five domains and three epistemic-marker classes. The article prints 90 questions but does not release participant choices or the marker assigned to each question. We preserve the printed questions and the reported class counts, then reconstruct assignments for proxy trials. This is a comparison with the published aggregate estimand, rather than an exact replay of the original trial sequence.

Let $d$ index the five domains, $e$ the three marker classes, $m$ a proxy model, $p$ a prompt protocol, and $i$ a trial. The response $Y_{mpei}$ is one for reliance and zero for independent lookup. Within a domain and marker class, $R_{mpde}$ is the mean of within-replicate reliance rates. The corresponding published human rate is $H_{de}$. We compare all 15 pairs $(R_{mpde},H_{de})$ and, for every marker class, all ten unordered domain pairs:
\begin{equation}
  \Delta_{mpe}^{a,b}=R_{mpae}-R_{mpbe},\qquad
  \Delta_{e}^{H,a,b}=H_{ae}-H_{be}.
  \label{eq:effect}
\end{equation}
The college-mathematics-minus-law effect is one of these pairs. We also score the strong-minus-weak marker effect within each domain. All domain pairs share cells, so their sign counts are descriptive rather than independent hypothesis tests. Absolute-rate and effect errors are arithmetic means of absolute proxy-minus-human differences, in percentage points. A zero proxy effect has no direction. In the thresholded all-pair count, both human and proxy effects must reach one percentage point; the human-resolved count instead requires a human working interval that excludes zero and a proxy effect of at least one point. Spearman correlation summarizes the ordering of the 15 rates; a constant proxy rate has no defined correlation. The zero-domain-effect rule sets each domain to the published mean within its marker class. We also display a constant at the overall human mean and a one-point mathematics-and-algebra increment on the marker means. All three rules are fitted to the scored human table. The STEM increment is a posthoc descriptive check on the resolved-pair structure, never a prospective predictor.

\subsection{Sampled binary replay}
\label{sec:method-core}

Each proxy sees a domain-labeled agent, a printed question, and an answer prefix containing a marker; the complete answer is withheld. The system instruction asks for a single \texttt{RELY} or \texttt{LOOK\_UP} decision. Each trial starts a fresh conversation. A seeded permutation assigns four questions per domain to strengtheners, ten to weakened strengtheners, and four to weakeners for each prompt replicate. The marker phrase is sampled within its class. The default protocol has no persona descriptor. The diverse protocol cycles evenly through eight descriptions crossing trivia familiarity, answer-checking habit, and chat-assistant use. We generate 48 prompt replicates under each protocol. A replicate is a repeated prompt construction and sample, not a heterogeneous human participant. The parser retains malformed text as missing and never coerces it to a decision.

\subsection{Counterbalanced probability control}
\label{sec:probability}

The control shows each printed question once under a representative phrase from each marker class and each of two option orders. In one order, A means rely; in the other, B means rely. The system instruction asks for A or B. We normalize the model's next-token logits over the A and B tokens and record the probability assigned to the rely option. The balanced rate is the arithmetic mean of the two order-specific probabilities, first within question and then within domain and marker class. This protocol has 540 prompt evaluations per model. Its order shift is the mean absolute difference between order-specific domain-by-marker rates. It measures a two-label conditional choice probability, not the probability of all possible free-form responses. Crossing marker classes within questions and using a fixed phrase also differ from the sampled replay, so the two protocols are separate measurements.

To separate a domain label from the question it accompanies, we cross every printed question with all five agent-domain labels while preserving its marker and both option orders. The matched crossover has 2,700 evaluations per model and is run for Qwen 7B, Qwen 72B AWQ, and Mistral Small. For each marker, a question-content marginal averages every question over the five labels before comparing its source domain with another source domain. An agent-label marginal averages each label over all 90 questions before comparing labels. The original question-label pairing is the diagonal of this design. These marginals are separate counterfactual contrasts, not additive components of the diagonal. For each source domain, we define congruence as reliance with the matching agent label minus mean reliance with the four other labels, averaging over its 18 questions and both option orders. Averaging congruence over all five source domains cancels additive label effects in this balanced design, leaving a question-domain-by-label interaction. Batching keeps all ten option-order and label variants of one question and marker together, so the within-question comparison shares a padding layout.

\subsection{Uncertainty and resource measures}
\label{sec:cost}

For the sampled replay we resample prompt replicates and printed questions independently within domain, recompute participant-first cell rates, and repeat this procedure 2,000 times. For the probability control we resample the 18 printed questions within each domain, preserving each question's two option orders and three marker classes. We use percentile intervals from these replicates for contrasts. Their coverage concerns these sampled proxy procedures; it does not absorb uncertainty from reconstructing the human marker assignments.

Rel-A.I. recruited 50 people for its domain setting and assigned four, ten, and four questions per class within each domain \citep{zhou2025rel}. Its aggregate table gives no participant records or covariance. For a transparent working comparison, we simulate independent binomial choices with the nominal per-cell decision counts and the published rates. One resample scored against the published table measures one-sided sampling error. Scoring two independent resamples against each other approximates a same-size human replication compared with a noisy reference. Both are conditional on the published rates as population probabilities and are not empirical human replications. Positive within-person correlation can widen cell intervals yet narrow within-person contrast intervals; its net effect cannot be identified from the table. We call a direction jointly resolved only when both the proxy interval and this human working interval strictly exclude zero. We also examine exchangeable-response correlations in Section~\ref{app:uncertainty}. No human-proxy difference is claimed to be statistically established from the published table alone.

Generation batches record input and output tokens and accelerator inference time. These quantities measure proxy computation; they exclude model loading and human-study administration. Rel-A.I. reports an hourly compensation minimum, but not total domain-study time or expenditure, so a dollar-per-effect comparison cannot be recovered from the published record.

\section{Experiments}
\label{sec:experiments}

\subsection{Setup}
\label{sec:setup}

We use the 90 questions printed in Rel-A.I. Table 7 and all 15 human domain-by-marker rates in its Table 4 \citep{zhou2025rel}. The sampled replay evaluates Qwen2.5-0.5B-Instruct, Qwen2.5-7B-Instruct, Mistral-7B-Instruct-v0.3, and Phi-3.5-mini-instruct at temperature 0.7. Each has 14,040 frozen trials across the default, diverse-persona, and greeting-probe protocols. The separate A/B probability control evaluates Qwen2.5-7B-Instruct, Qwen2.5-32B-Instruct, Qwen2.5-72B-Instruct-AWQ, and Mistral-Small-Instruct-2409. Three of these models also enter the matched domain-label crossover. No prompt supplies human rates, hypotheses, or answer keys. Tables show point estimates; resampling endpoints are in Section~\ref{app:uncertainty}.

\subsection{Absolute rates and sampled decisions}
\label{sec:main-results}

\begin{figure}[H]\centering
  \includegraphics[width=\columnwidth]{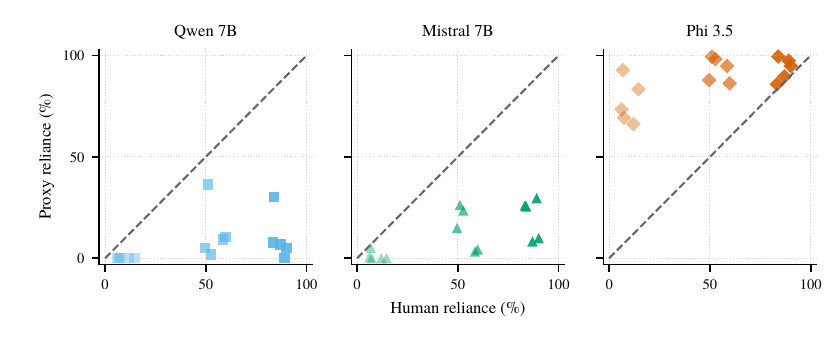}
  \caption{\textbf{Sampled proxies depart from human absolute rates.} Each panel places the 15 default-prompt rates of one nonconstant proxy against published human rates. The dashed diagonal is equality. Symbol shape and color identify the model; lighter symbols indicate weaker marker classes. Qwen 0.5B's constant-zero rates remain in Table~\ref{tab:allrates}.}
  \label{fig:allrates}
\end{figure}

Table~\ref{tab:allrates} reports every sampled-replay rate beside its human reference. The proxies occupy different response regimes. Qwen 0.5B always looks up the answer in the domain trials. Qwen 7B and Mistral 7B are near the floor for weak markers, while Phi 3.5 mini is near the ceiling for strong and moderate markers. Figure~\ref{fig:allrates} shows these level errors across all 15 cells. A saturated decision rule has little room to express a domain shift.

\begin{table*}[t]
\centering\small
\caption{Reliance rate (\%) in every published domain--marker cell. Human values are from Rel-A.I. Table 4; proxy values are default-prompt estimates on reconstructed assignments.}
\label{tab:allrates}
\begin{tabular}{llrrrrr}
\toprule
Domain & Marker & Human & Qwen 0.5B & Qwen 7B & Mistral 7B & Phi 3.5 \\
\midrule
College math & Strong & 90.1 & 0.0 & 5.2 & 9.9 & 94.8 \\
College math & Moderate & 58.5 & 0.0 & 9.2 & 3.3 & 94.8 \\
College math & Weak & 14.6 & 0.0 & 0.0 & 0.0 & 83.3 \\
\addlinespace
Abstract algebra & Strong & 87.0 & 0.0 & 6.8 & 8.3 & 89.6 \\
Abstract algebra & Moderate & 59.8 & 0.0 & 10.6 & 4.4 & 86.3 \\
Abstract algebra & Weak & 12.0 & 0.0 & 0.0 & 0.0 & 66.1 \\
\addlinespace
Philosophy & Strong & 89.1 & 0.0 & 0.0 & 29.7 & 97.4 \\
Philosophy & Moderate & 52.7 & 0.0 & 1.9 & 23.5 & 98.1 \\
Philosophy & Weak & 6.2 & 0.0 & 0.0 & 0.5 & 73.4 \\
\addlinespace
World religion & Strong & 83.9 & 0.0 & 30.2 & 25.5 & 99.5 \\
World religion & Moderate & 51.0 & 0.0 & 36.5 & 26.2 & 99.4 \\
World religion & Weak & 6.8 & 0.0 & 0.0 & 5.2 & 92.7 \\
\addlinespace
Law & Strong & 83.3 & 0.0 & 7.8 & 26.0 & 85.8 \\
Law & Moderate & 49.6 & 0.0 & 5.2 & 15.0 & 87.8 \\
Law & Weak & 7.3 & 0.0 & 0.0 & 0.0 & 69.3 \\
\bottomrule
\end{tabular}
\end{table*}

\begin{figure}[H]
  \centering
  \includegraphics[width=\columnwidth]{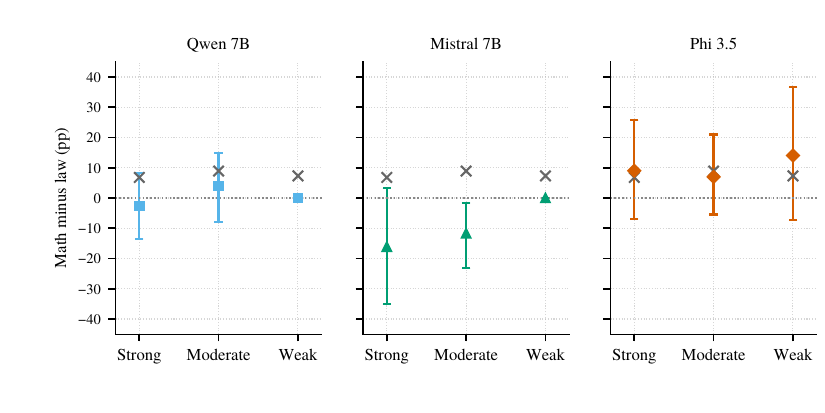}
  \caption{\textbf{The sampled mathematics-minus-law effects remain uncertain.} Colored points show default-prompt proxy estimates and vertical bars show percentile intervals from resampling prompt replicates and questions. Grey crosses are published human effects. The human working intervals are evaluated separately because participant records are unavailable.}
  \label{fig:uncertainty}
\end{figure}

Table~\ref{tab:allsummary} reports the sampled models beside a plug-in same-size human resample. Its expected domain-effect error is 2.5 points against the published rates; a two-sample human comparison has 3.5 points of expected error, while the zero-domain-effect rule scores 4.6. No sampled model improves on that rule. Among human-resolved pairs, Mistral 7B has five jointly resolved proxy effects and matches none; all five are moderate-marker contrasts listed in Section~\ref{app:uncertainty}. Phi's three mathematics-minus-law point estimates share the human direction, but each corresponding proxy interval includes zero (Figure~\ref{fig:uncertainty}).

\begin{table*}[t]
\centering\small
\caption{Default-prompt agreement across all 15 rates and 30 domain-pair contrasts. Errors are percentage points. H hits are matched human-resolved pairs with a proxy effect of at least 1 pp. Joint res. counts pairs for which both working intervals exclude zero; joint hits counts their matching directions. Human resample is an independent-binomial plug-in reference; oracle rules use the scored human table.}
\label{tab:allsummary}
\begin{tabular}{lrrrrrr}
\toprule
Proxy & Rate MAE & Rate $\rho$ & Effect MAE & H hits & Joint res. & Joint hits \\
\midrule
Qwen 0.5B & 50.1 & -- & 4.6 & 0 & 0 & 0 \\
Qwen 7B & 42.6 & 0.51 & 12.0 & 3 & 3 & 1 \\
Mistral 7B & 38.3 & 0.61 & 12.4 & 0 & 5 & 0 \\
Phi 3.5 & 37.8 & 0.55 & 10.3 & 3 & 2 & 0 \\
Oracle overall mean & 27.2 & -- & 4.6 & 0 & -- & -- \\
Oracle marker mean & 3.2 & 0.94 & 4.6 & 0 & -- & -- \\
Posthoc STEM rule & 2.8 & 0.98 & 4.0 & 8 & -- & -- \\
Human resample & 1.8 & -- & 2.5 & -- & -- & -- \\
\bottomrule
\end{tabular}
\end{table*}

\subsection{Does a larger, counterbalanced proxy improve agreement?}
\label{sec:choice-results}

Table~\ref{tab:choicecells} makes the probability control's 15 cells inspectable for all four models. Mistral Small has the lowest absolute-rate error among these controls, 22.5 percentage points; Qwen 72B has 30.9 (Table~\ref{tab:choice}). Their domain-effect errors are 8.3 and 19.6 points, respectively, against 3.5 for a two-sample plug-in human comparison and 4.6 for the zero-domain-effect rule. Qwen 72B has rate rank correlation 0.87 yet large level and effect errors. The table-fitted marker and STEM rules are descriptive oracles, not prospective competitors. These comparisons are conditional on this prompt and the batch layout; they do not identify model-intrinsic validity.

\begin{table*}[t]
\centering\footnotesize
\caption{Human and counterbalanced A/B reliance rates (\%). Every value is a point estimate. S, M, W denote strong, moderate, and weak epistemic markers.}
\label{tab:choicecells}
\begin{tabular}{llrrrrr}
\toprule
Domain & Marker & Human & Qwen 7B & Qwen 32B & Qwen 72B & Mistral Small \\
\midrule
College math & S & 90.1 & 37.1 & 74.0 & 55.2 & 43.2 \\
College math & M & 58.5 & 18.2 & 43.3 & 48.8 & 35.0 \\
College math & W & 14.6 & 1.4 & 0.0 & 0.9 & 13.5 \\
\addlinespace
Abstract algebra & S & 87.0 & 21.7 & 59.2 & 62.8 & 36.4 \\
Abstract algebra & M & 59.8 & 16.1 & 15.6 & 51.3 & 27.8 \\
Abstract algebra & W & 12.0 & 2.9 & 0.0 & 0.6 & 11.0 \\
\addlinespace
Philosophy & S & 89.1 & 0.0 & 0.1 & 13.1 & 52.6 \\
Philosophy & M & 52.7 & 1.1 & 0.0 & 6.2 & 45.1 \\
Philosophy & W & 6.2 & 0.0 & 0.0 & 0.0 & 14.5 \\
\addlinespace
World religion & S & 83.9 & 0.4 & 21.8 & 42.7 & 49.1 \\
World religion & M & 51.0 & 0.1 & 0.0 & 5.3 & 34.4 \\
World religion & W & 6.8 & 0.0 & 0.0 & 0.0 & 9.1 \\
\addlinespace
Law & S & 83.3 & 0.0 & 5.5 & 1.8 & 33.4 \\
Law & M & 49.6 & 0.0 & 0.0 & 0.3 & 23.7 \\
Law & W & 7.3 & 0.7 & 0.0 & 0.0 & 7.3 \\
\bottomrule
\end{tabular}
\end{table*}

\begin{table*}[t]
\centering\footnotesize
\caption{A/B control, human resampling reference, and table-fitted diagnostic rules. Errors are percentage points; marker error scores five strong-minus-weak contrasts. The 1-pp sign count requires both human and proxy effects to reach that magnitude. H-res. hits counts matched human-resolved pairs out of eight. The human row averages independent-binomial samples using published rates as plug-in probabilities, so its counts can be fractional. Oracle and posthoc rules are descriptive, not deployable predictors.}
\label{tab:choice}
\begin{tabular}{lrrrrrr}
\toprule
Protocol & Rate MAE & Rate $\rho$ & Domain MAE & Marker MAE & $\geq$1 pp & H-res. hits \\
\midrule
Oracle overall & 27.2 & -- & 4.6 & 77.3 & 0 & 0 \\
Oracle marker & 3.2 & 0.94 & 4.6 & 2.2 & 0 & 0 \\
Posthoc STEM & 2.8 & 0.98 & 4.0 & 2.2 & 17 & 8 \\
Qwen 7B & 43.5 & 0.40 & 9.1 & 66.5 & 18 & 7 \\
Qwen 32B & 35.5 & 0.88 & 20.0 & 45.2 & 12 & 5 \\
Qwen 72B AWQ & 30.9 & 0.87 & 19.6 & 42.5 & 15 & 5 \\
Mistral Small & 22.5 & 0.84 & 8.3 & 45.5 & 16 & 4 \\
Human resample & 1.8 & -- & 2.5 & 2.5 & 22.6 & 7.9 \\
\bottomrule
\end{tabular}
\end{table*}

Only eight human pairs have discrete-binomial working intervals that strictly exclude zero. The plug-in human resample averages 22.6 thresholded sign matches and 7.9 hits on these pairs (Table~\ref{tab:choice}). All eight compare mathematics or algebra with a humanities domain. Qwen 7B matches seven after a one-point proxy-effect threshold, while the posthoc STEM rule matches eight by construction. A common permutation of the five domain labels matches or exceeds Qwen 7B's seven hits in 12 of 120 assignments. This test has little resolution beyond the STEM-versus-humanities partition. A normal working model resolves ten pairs at zero within-person correlation and fourteen at correlation 0.3; the STEM rule then matches ten and thirteen, respectively (Section~\ref{app:uncertainty}). The discrete-versus-normal difference concerns boundary pairs whose rounded effects sit near zero. The 30 pairs share only 15 cells and are not independent confirmations.

The question-resampling analysis preserves both option orders within each item. With the human table fixed, each control's domain-effect-error interval stays above 4.6 points; Section~\ref{app:uncertainty} prints those endpoints. Counterbalancing averages the two orders but does not remove wording sensitivity. Appendix~\ref{app:order} reports the observed shift, and the experiment artifact retains each order-specific cell rate.

\subsection{Does the domain cue survive matched questions?}
\label{sec:crossover}

The matched crossover separates agent-label effects from differences among the domain-specific question sets. Appendix~\ref{app:matrix} shows mathematics-minus-law content and label marginals in Figure~\ref{fig:crossover}. Qwen 72B's strong-marker label marginal is 21.1 points and Mistral Small's is $-0.4$; neither has a matched human label-only counterpart. The comparable diagonal contrasts are 53.6 and 9.8 points against the published human 6.8 points. Qwen 7B's label-only domain-effect error is 3.4 points, below the 4.6-point zero-effect rule and within the one-sided human-resample range. Its question-bootstrap interval remains below the zero-effect error (Section~\ref{app:uncertainty}). The diagonal, which matches the human study's coupled question and label assignment, has 9.6-point error. Label-only agreement therefore does not validate the deployed diagonal protocol.

\begin{table*}[t]
\centering\small
\caption{Matched crossover point estimates (pp). Domain MAE compares the human 30-pair table with diagonal, content, or label marginals. Congruence is reliance under the matching agent label minus the mean of four other labels, averaged over the five source-question domains. Human and zero-effect references apply only to the domain-MAE columns.}
\label{tab:crossover}
\begin{tabular}{lrrrrrr}
\toprule
Proxy & Diag. & Content & Label & Cong. S & Cong. M & Cong. W \\
\midrule
Qwen 7B & 9.6 & 5.2 & 3.4 & 6.9 & 4.2 & 0.3 \\
Qwen 72B AWQ & 19.7 & 10.2 & 5.6 & 23.7 & 15.0 & 0.2 \\
Mistral Small & 8.3 & 6.1 & 5.3 & 13.3 & 9.6 & 2.4 \\
\midrule
Zero domain effect & 4.6 & 4.6 & 4.6 & -- & -- & -- \\
Two-sample human & 3.5 & -- & -- & -- & -- & -- \\
\bottomrule
\end{tabular}
\end{table*}

\begin{figure}[H]
  \centering
  \includegraphics[width=\columnwidth]{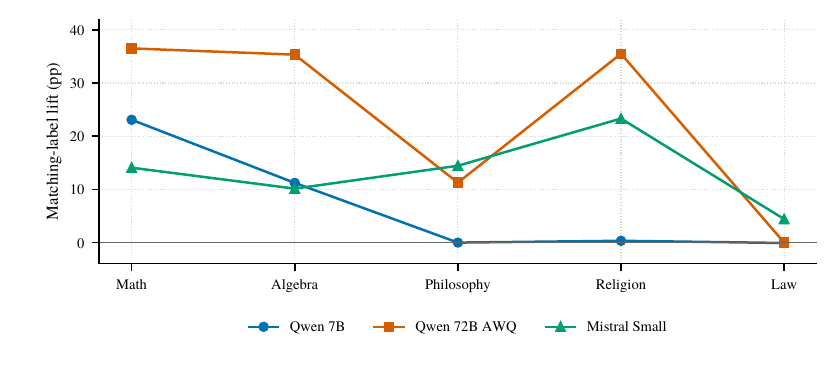}
  \caption{\textbf{The matching-label effect persists when the global label marginal is small.} Strong-marker matching-label lift compares each question's own agent-domain label with the mean of its four other labels, within source domain. Points are measured domain means from the matched crossover; zero denotes no within-domain label preference.}
  \label{fig:congruence}
\end{figure}

Table~\ref{tab:crossover} also reports congruence, a matched label effect within each question domain. Mistral Small has a 13.3-point mean strong-marker matching-label lift despite its near-zero mathematics-minus-law label marginal. Figure~\ref{fig:congruence} shows the lift for every source domain. This is a question-by-label interaction: averaging labels across unrelated questions can cancel substantial within-domain responses. The crossover identifies a response to the printed cue, but no answer key or independent confidence measure is available to determine how much of the content marginal tracks question knowledge.

\subsection{Persona and resource checks}
\label{sec:ablations}

The diverse-persona prompt changes the sampled models' levels and domain contrasts without making them human-calibrated. Phi 3.5 mini's absolute-rate error falls from 37.8 to 20.0 points, while its domain-effect error rises from 10.3 to 17.2 points. The complete diverse-persona rates are in Appendix~\ref{app:persona}; class-specific contrasts follow from those rates and Table~\ref{tab:allrates}. Prompt variation can improve one validity measure while worsening another.

The greeting probe uses printed domain questions rather than the unreleased human greeting trials and is diagnostic only (Appendix~\ref{app:source}). Sampled parse failures number 0, 0, 1, and 15 for Qwen 0.5B, Qwen 7B, Mistral 7B, and Phi 3.5 mini, respectively, among 14,040 trials per model. They remain in the raw records and are excluded only from rate denominators. The 540-prompt A/B controls consumed 63,294 input tokens for each Qwen model and 66,018 for Mistral Small. Model loading and the unreleased human-study expenditure prevent a matched dollar-per-effect comparison.

\section{Uncertainty details}
\label{app:uncertainty}

The human working approximation uses nominal counts of 200 decisions for a strong or weak marker cell and 500 for a moderate-marker cell. Eight of the 30 domain contrasts have discrete-binomial percentile intervals strictly excluding zero. For mathematics minus law, its percentile intervals in percentage points are $[0.0,13.5]$, $[2.6,15.4]$, and $[1.0,13.5]$ for strong, moderate, and weak markers. These are not participant-level intervals. The corresponding two-way proxy intervals for Phi 3.5 mini under the default prompt are $[-6.8,25.7]$, $[-5.4,21.0]$, and $[-7.3,36.6]$. The probability-control question intervals for Qwen 32B are $[50.9,84.5]$, $[24.4,63.1]$, and $[0.0,0.0]$ for the same ordered effects; the last pair is numerically degenerate at displayed precision and does not establish certainty. A normal sensitivity model at zero within-person correlation resolves ten pairs. It adds strong-marker mathematics--law and weak-marker algebra--philosophy relative to the discrete percentile rule. This difference arises from a normal critical value applied to a discrete, boundary-sensitive binomial statistic. At correlation 0.3, fourteen pairs resolve; one added within-humanities pair breaks the posthoc STEM rule's perfect count. The model's cross-domain covariance partly offsets increased cell variance. Table~\ref{tab:pairs} marks the membership of all three resolved sets.

For the crossover label-only domain-effect error, question-bootstrap intervals are $[3.1,4.0]$, $[4.5,6.7]$, and $[4.7,6.0]$ points for Qwen 7B, Qwen 72B, and Mistral Small. Qwen 7B's label marginal falls below the 4.6-point zero-effect rule in 99.95\% of bootstrap replicates. The one-sided human-resample range is $[1.5,3.6]$ points; its overlap with Qwen 7B's interval does not establish equivalence, and the human label-only estimand is unobserved. The two-sample resampling calculation gives 2.5-point rate error and 3.5-point domain-effect error in expectation.

In that model order, diagonal domain-error intervals are $[6.5,13.7]$, $[16.5,23.6]$, and $[7.0,9.9]$ points; content-marginal intervals are $[3.9,8.0]$, $[7.9,13.0]$, and $[4.9,7.7]$. Strong-marker mean congruence intervals are $[4.2,10.0]$, $[20.4,27.0]$, and $[11.8,14.9]$ points. These intervals resample questions within source domain, preserving each question's labels and option orders.

Mistral 7B's five jointly resolved sampled effects are moderate-marker contrasts: mathematics against world religion and law, and abstract algebra against philosophy, world religion, and law. Their proxy directions oppose the human directions. Full cell and pairwise endpoints for every model remain in the machine-readable audit.

\section{Conclusion}
\label{sec:conclusion}

Across the released Rel-A.I. domain table, the tested sampled and counterbalanced proxies have large reliance-level and diagonal effect errors. The matched crossover shows that agent-domain labels have question-dependent effects: Mistral Small's small global math-minus-law label marginal coexists with a substantial matching-label lift. Question-set differences remain after averaging over labels, but the models' own answer accuracy is unknown. Identical-prompt probabilities also vary with inference layout. The human aggregate resolves few domain contrasts and leaves within-person covariance unavailable. This single-study audit does not establish proxy-based directional screening or replacement of human reliance measurements under the tested protocols.

\section*{Limitations}
The human reference contains 15 rates from one study with 50 people, rather than released trial-level decisions. The binomial working intervals use nominal decision counts and ignore dependence among decisions by one person. The bias can point either way for within-person contrasts; these intervals cannot substitute for a participant-level bootstrap or the original study's covariance. Only eight of the 30 domain pairs resolve under this working model, limiting any directional validity claim against this table. A plug-in independent human resample scored against the published table measures one-sided noise; the two-sample comparison is the closer approximation to a new human replication. Both omit within-person dependence. The 30 domain-pair effects share the same 15 cells and are not independent confirmations.

The printed questions are preserved, but their marker assignments are reconstructed. Different assignments, interface timing, or participant instructions could alter the comparison. The five domains contain distinct question sets, so domain differences can reflect question content, difficulty, or model knowledge alongside perceived domain competence. The sampled replay's prompt replicates differ by permutations and random generation; without persona descriptors, their spread is sampling and item-assignment variation, not human heterogeneity. One generator seed was used, so the bootstrap does not estimate between-seed generation variation. The diverse descriptors are a prompt intervention and are not sampled from a human demographic distribution. An affine calibration fitted and scored on the same 15 human cells would be in-sample; no separate human domain table is available for a prospective calibration test.

The probability control uses one marker phrase per class and crosses all classes with every question. It therefore answers a related sensitivity question under a cleaner within-item design, not an exact reanalysis of the original human trials. A/B-normalized next-token probabilities depend on label wording and inference-batch layout. The identical diagonal prompts in the four-item control and ten-item crossover have cellwise mean absolute differences of 0.36, 0.08, and 0.07 points for Qwen 7B, Qwen 72B, and Mistral Small; their largest cell differences are 1.88, 0.33, and 0.37 points. At the individual prompt level the largest differences are 21.51, 2.15, and 2.60 points. This is an unresolved implementation sensitivity, so neither cross-batch difference is interpreted as a behavioral effect. Within-crossover label comparisons share a batch layout, although that does not establish batch-size-one invariance. Label format, marker phrase, question assignment, and sampling change together between the sampled and probability protocols, so their difference has no identified cause. The largest evaluated proxy has 72B parameters in four-bit AWQ form, but no frontier API model was tested. Quantization may affect its responses. The single English, closed-form reliance task cannot establish validity in longer conversations, other languages, or consequential decisions. No additional human reliance dataset with released compatible trials was used. The crossover isolates the printed domain cue while holding the question fixed; no answer key was released, so it does not measure the proxy's own question accuracy.

The greeting probe reuses domain questions because the human greeting-study item assignments were not released. Its result cannot be scored against the human greeting effect. The paper's evidence therefore concerns domain-conditioned reliance; it does not validate greeting or prior-confidence effect sizes. Rel-A.I. reports an hourly participant compensation minimum but no total domain-study time or expenditure, preventing a like-for-like dollar cost comparison.

\section*{Use of AI Assistants}
The experiments and the manuscript were produced with substantial assistance from LLM-based research agents. The authors take full responsibility for the content of this paper.

\bibliography{main}

\appendix
\section{Additional interval tables}
\paragraph{Resampling endpoints.}
The plug-in human resample gives rate MAE 1.8 pp with a 95\% Monte Carlo range [1.1, 2.5], and domain-effect MAE 2.5 pp with range [1.5, 3.6]. The reference is conditional on independent binomial responses and published rates as population probabilities.
Question-resampling 95\% intervals for A/B domain-effect MAE are Qwen 7B [6.1, 13.1], Qwen 32B [15.5, 24.6], Qwen 72B AWQ [16.5, 23.5], Mistral Small [6.9, 9.9] pp, with the human table held fixed.
Paired-question 95\% intervals for the strong-marker mathematics-minus-law agent-label effect are Qwen 7B [8.5, 16.0], Qwen 72B AWQ [17.8, 24.5], Mistral Small [-2.4, 1.3] pp.
The exact common-domain-label permutation tail on the discrete-binomial human-resolved set is Qwen 7B 12/120, Qwen 32B 12/120, Qwen 72B AWQ 12/120, Mistral Small 62/120. The test preserves the five-domain partition and has limited power when the resolved signs mostly encode a STEM-versus-humanities split.
The statistic is the number of thresholded matching directions on the fixed human-resolved set after a common permutation of the five printed domain labels. Observed counts differ across the Qwen models, but each has the same tail count because the permutations reproduce the same coarse STEM-versus-humanities pattern.

\begin{table*}[t]
\centering\footnotesize
\caption{Matched human-resolved directions under three working interval choices. B uses the discrete binomial bootstrap; zero and 0.3 are within-person correlation settings in the normal sensitivity model. The STEM rule is fitted to the human table. Each entry is a matched-pair count; all proxy effects must reach 1 pp.}
\label{tab:sensitivity}
\begin{tabular}{lrrrrrr}
\toprule
Human model & Pairs & STEM & Qwen 7B & Qwen 32B & Qwen 72B & Mistral Small \\
\midrule
Discrete B & 8 & 8 & 7 & 5 & 5 & 4 \\
Corr. 0 & 10 & 10 & 9 & 6 & 6 & 5 \\
Corr. 0.3 & 14 & 13 & 12 & 8 & 9 & 7 \\
\bottomrule
\end{tabular}
\end{table*}

\section{Reproducibility and compute}
\label{app:repro}

The artifact retains questions, human rates, trial manifests, raw model outputs, aggregates, resampling intervals, and exporters. The sampled runs each include 14,040 frozen trials, with 48 prompt replicates per persona protocol and ten replicates in the crossed greeting probe. All sampled models used bfloat16 weights, temperature 0.7, generator seed 82717, and a 12-token output cap on an H100 accelerator. The probability control evaluates every printed question under three representative markers and both option orders. The matched crossover adds all five domain labels for every question and marker, with ten variants of each question-marker pair in one batch. Qwen 7B, Qwen 32B, and Mistral Small use bfloat16 weights; Qwen 72B uses a four-bit AWQ checkpoint. Each prompt is an independent chat turn. The exact model revisions, artifact checksums, and hardware records are retained in the artifact's experiment-details record.

The sampled parser accepts a leading \texttt{RELY} or \texttt{LOOK\_UP} token. Responses without either leading token remain in the raw records and are omitted only from rate denominators. The probability control normalizes the two next-token logits for A and B; it does not generate free-form continuations. Cell rates are computed before domain contrasts, and the two option orders are averaged within each question. The uncertainty analysis uses 2,000 resamples with fixed seeds, resampling both prompt replicate and question for sampled decisions and resampling question for the deterministic probability control.

\section{Evidence-source boundary and cue probe}
\label{app:source}

The 90 questions and human rates come from Rel-A.I. Tables 7 and 4 \citep{zhou2025rel}. Its reported per-class question counts determine the sampled permutations; the probability control crosses every question with a representative phrase from each class. Neither protocol reconstructs participant histories.

The cue probe crosses ``I'm happy to help!'' with three markers on the printed questions. Its paired shifts are 0.0, 3.8, 9.2, and 3.0 percentage points for Qwen 0.5B, Qwen 7B, Mistral 7B, and Phi 3.5 mini. Unreleased human greeting-trial assignments preclude a matched human comparison.

\section{Choice-order sensitivity}
\label{app:order}
The A/B protocol averages both option orders. Mean absolute cellwise shifts before averaging are 4.1, 7.1, 8.2, and 7.6 points for Qwen 7B, Qwen 32B, Qwen 72B, and Mistral Small. The corresponding maximum single-cell shifts are 21.4, 24.8, 22.6, and 12.5 points. These measurements isolate option-order sensitivity within the probability protocol. Inference times are not a controlled cost comparison because loading, quantization, and CPU offload differ among models.

\section{Pairwise domain audit}
\label{app:pairs}

Table~\ref{tab:pairs} gives all 30 point-estimate contrasts and three human-resolution flags. An endpoint at zero does not resolve a direction. Full-precision effects and intervals are retained in the experiment artifact.
\begingroup\scriptsize\setlength{\tabcolsep}{3pt}
\begin{table*}[t]
\centering
\caption{Within-marker domain effects in percentage points. B denotes the discrete-binomial working interval; 0 and .3 denote normal within-person-correlation sensitivity settings. Y marks a human effect whose interval excludes zero. Proxy values are point estimates; full intervals remain with experiment details.}\label{tab:pairs}
\begin{tabular}{llrcccrrrrr}
\toprule Marker & Pair & Human & B & 0 & .3 & Phi & Q7 & Q32 & Q72 & Mistral \\
\midrule
S & Math--Alg. & 3.1 & -- & -- & -- & 5.2 & 15.4 & 14.8 & -7.6 & 6.8 \\
S & Math--Phil. & 1.0 & -- & -- & -- & -2.6 & 37.1 & 73.9 & 42.1 & -9.4 \\
S & Math--Rel. & 6.2 & -- & -- & Y & -4.7 & 36.6 & 52.2 & 12.5 & -5.9 \\
S & Math--Law & 6.8 & -- & Y & Y & 9.0 & 37.1 & 68.5 & 53.4 & 9.8 \\
S & Alg.--Phil. & -2.1 & -- & -- & -- & -7.8 & 21.7 & 59.1 & 49.7 & -16.1 \\
S & Alg.--Rel. & 3.1 & -- & -- & -- & -9.9 & 21.3 & 37.4 & 20.1 & -12.6 \\
S & Alg.--Law & 3.7 & -- & -- & -- & 3.8 & 21.7 & 53.6 & 61.0 & 3.0 \\
S & Phil.--Rel. & 5.2 & -- & -- & -- & -2.1 & -0.4 & -21.7 & -29.6 & 3.5 \\
S & Phil.--Law & 5.8 & -- & -- & Y & 11.6 & 0.0 & -5.4 & 11.3 & 19.1 \\
S & Rel.--Law & 0.6 & -- & -- & -- & 13.7 & 0.4 & 16.3 & 40.9 & 15.6 \\
\addlinespace
M & Math--Alg. & -1.3 & -- & -- & -- & 8.5 & 2.1 & 27.7 & -2.5 & 7.2 \\
M & Math--Phil. & 5.8 & -- & -- & Y & -3.3 & 17.1 & 43.3 & 42.6 & -10.1 \\
M & Math--Rel. & 7.5 & Y & Y & Y & -4.6 & 18.1 & 43.3 & 43.4 & 0.6 \\
M & Math--Law & 8.9 & Y & Y & Y & 7.0 & 18.2 & 43.3 & 48.5 & 11.3 \\
M & Alg.--Phil. & 7.1 & Y & Y & Y & -11.9 & 15.0 & 15.6 & 45.0 & -17.3 \\
M & Alg.--Rel. & 8.8 & Y & Y & Y & -13.1 & 15.9 & 15.6 & 45.9 & -6.6 \\
M & Alg.--Law & 10.2 & Y & Y & Y & -1.5 & 16.1 & 15.6 & 51.0 & 4.1 \\
M & Phil.--Rel. & 1.7 & -- & -- & -- & -1.2 & 1.0 & 0.0 & 0.9 & 10.7 \\
M & Phil.--Law & 3.1 & -- & -- & -- & 10.3 & 1.1 & 0.0 & 6.0 & 21.4 \\
M & Rel.--Law & 1.4 & -- & -- & -- & 11.6 & 0.1 & 0.0 & 5.1 & 10.7 \\
\addlinespace
W & Math--Alg. & 2.6 & -- & -- & -- & 17.2 & -1.5 & 0.0 & 0.2 & 2.5 \\
W & Math--Phil. & 8.4 & Y & Y & Y & 9.9 & 1.4 & 0.0 & 0.8 & -1.0 \\
W & Math--Rel. & 7.8 & Y & Y & Y & -9.4 & 1.3 & 0.0 & 0.9 & 4.4 \\
W & Math--Law & 7.3 & Y & Y & Y & 14.1 & 0.6 & 0.0 & 0.9 & 6.2 \\
W & Alg.--Phil. & 5.8 & -- & Y & Y & -7.3 & 2.9 & 0.0 & 0.6 & -3.5 \\
W & Alg.--Rel. & 5.2 & -- & -- & Y & -26.6 & 2.9 & 0.0 & 0.6 & 1.9 \\
W & Alg.--Law & 4.7 & -- & -- & -- & -3.1 & 2.1 & 0.0 & 0.6 & 3.7 \\
W & Phil.--Rel. & -0.6 & -- & -- & -- & -19.3 & 0.0 & 0.0 & 0.0 & 5.3 \\
W & Phil.--Law & -1.1 & -- & -- & -- & 4.2 & -0.7 & 0.0 & 0.0 & 7.2 \\
W & Rel.--Law & -0.5 & -- & -- & -- & 23.4 & -0.7 & 0.0 & 0.0 & 1.8 \\
\addlinespace
\bottomrule
\end{tabular}
\end{table*}
\endgroup

\section{Persona rate audit}
\label{app:persona}
\begin{table*}[t]
\centering\footnotesize
\caption{Diverse-persona sampled reliance rates (\%) for every domain and marker class. Point estimates are shown; the corresponding default rates appear in Table~\ref{tab:allrates}.}
\label{tab:persona_audit}
\begin{tabular}{llrrrr}
\toprule
Domain & Marker & Qwen 0.5B & Qwen 7B & Mistral 7B & Phi 3.5 \\
\midrule
College math & S & 0.0 & 4.2 & 8.3 & 80.2 \\
College math & M & 0.0 & 5.6 & 5.0 & 77.1 \\
College math & W & 0.0 & 0.0 & 0.5 & 44.3 \\
\addlinespace
Abstract algebra & S & 0.0 & 1.0 & 9.4 & 66.1 \\
Abstract algebra & M & 0.0 & 3.1 & 5.2 & 65.2 \\
Abstract algebra & W & 0.0 & 0.0 & 2.1 & 20.8 \\
\addlinespace
Philosophy & S & 0.0 & 0.0 & 35.4 & 88.0 \\
Philosophy & M & 0.0 & 0.4 & 24.2 & 88.5 \\
Philosophy & W & 0.0 & 0.0 & 5.2 & 19.3 \\
\addlinespace
World religion & S & 0.0 & 23.4 & 30.7 & 95.3 \\
World religion & M & 0.0 & 27.3 & 29.8 & 93.3 \\
World religion & W & 0.0 & 0.0 & 7.3 & 54.2 \\
\addlinespace
Law & S & 0.0 & 5.2 & 20.3 & 79.7 \\
Law & M & 0.0 & 3.3 & 14.4 & 72.0 \\
Law & W & 0.0 & 0.0 & 1.6 & 36.5 \\
\bottomrule
\end{tabular}
\end{table*}

\section{Strong-marker crossover matrix}
\label{app:matrix}

Table~\ref{tab:matrix} exposes the complete source-question by agent-label rates for the strong-marker crossover. The corresponding moderate- and weak-marker matrices are retained with the machine-readable results. Congruence in Table~\ref{tab:crossover} subtracts each row's four off-diagonal cells from its diagonal cell before averaging source domains.
\begin{table*}[t]
\centering\footnotesize
\caption{Strong-marker crossover matrix: source question domain by printed agent-domain label. Every cell is the balanced reliance rate (\%). Matching labels are on the diagonal.}
\label{tab:matrix}
\begin{tabular}{llrrrrr}
\toprule
Proxy & Questions & Math & Algebra & Philosophy & Religion & Law \\
\midrule
Qwen 7B & Math & 39.0 & 19.6 & 22.0 & 12.1 & 9.9 \\
 & Algebra & 32.4 & 21.8 & 5.4 & 3.0 & 1.3 \\
 & Philosophy & 0.0 & 0.0 & 0.0 & 0.0 & 0.0 \\
 & Religion & 0.1 & 0.0 & 0.2 & 0.4 & 0.0 \\
 & Law & 0.1 & 0.0 & 0.2 & 0.0 & 0.0 \\
\addlinespace
Qwen 72B AWQ & Math & 55.4 & 44.0 & 16.8 & 8.2 & 6.7 \\
 & Algebra & 65.8 & 62.9 & 23.0 & 11.9 & 9.5 \\
 & Philosophy & 1.9 & 1.0 & 13.1 & 4.2 & 0.3 \\
 & Religion & 2.5 & 3.1 & 21.0 & 42.8 & 2.8 \\
 & Law & 0.8 & 0.4 & 5.1 & 0.7 & 1.8 \\
\addlinespace
Mistral Small & Math & 43.2 & 36.0 & 29.4 & 20.2 & 30.8 \\
 & Algebra & 37.5 & 36.4 & 23.1 & 17.7 & 26.9 \\
 & Philosophy & 34.9 & 30.0 & 52.2 & 44.2 & 41.9 \\
 & Religion & 17.6 & 16.4 & 37.8 & 49.0 & 31.0 \\
 & Law & 28.8 & 22.7 & 36.1 & 28.5 & 33.4 \\
\bottomrule
\end{tabular}
\end{table*}

\begin{figure}[H]
  \centering
  \includegraphics[width=\columnwidth]{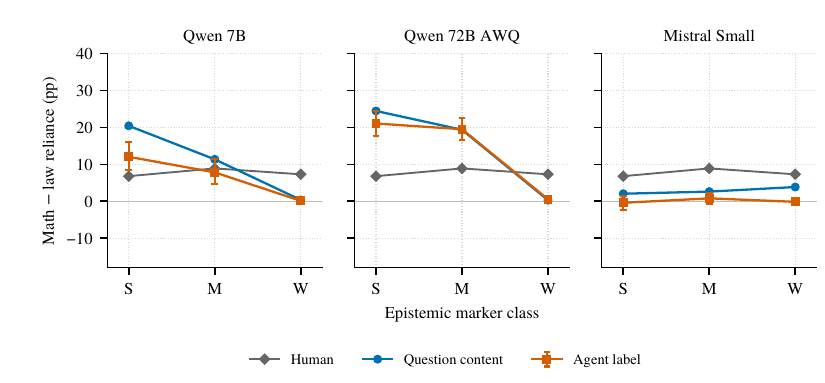}
  \caption{\textbf{Content and label marginals differ across models.} Mathematics-minus-law effects are shown for strong (S), moderate (M), and weak (W) markers. Blue circles compare source question sets after averaging over agent labels; vermilion squares compare agent labels after averaging over all questions. Grey diamonds are published human diagonal effects and are not matched label-only effects. Vertical bars are percentile intervals from resampling questions within source domain.}
  \label{fig:crossover}
\end{figure}

\section{Prompt construction}
\label{app:prompt}

The sampled prompt requests \texttt{RELY} or \texttt{LOOK\_UP} after a domain-labeled question and hidden-answer prefix; the A/B prompt requests one option letter. Fixed strong, moderate, and weak prefixes are ``I'm certain it's'', ``I think it's'', and ``I'm not sure it's''. Diverse personas cross trivia familiarity, answer-checking habit, and chat-assistant use. Exact prompts accompany the trial manifest.

\section{Computational resources}
The documented probability and label-crossover analyses used 0.122 H100 GPU-hours. Earlier sampled-model experiments and other probability controls are outside this measured subset. A complete GPU-hour budget for the study is unavailable.

\end{document}